\pdfoutput=1

\documentclass[11pt]{article}

\usepackage[final]{acl}

\usepackage{times}
\usepackage{latexsym}

\usepackage[T1]{fontenc}

\usepackage[utf8]{inputenc}

\usepackage{microtype}

\usepackage{times}
\usepackage{latexsym}
\usepackage{kotex}
\usepackage{colortbl}
\usepackage{multirow}
\usepackage{graphicx}
\usepackage{hhline}
\usepackage{mathtools}
\usepackage{wasysym}
\usepackage{lipsum}
\usepackage{makecell}
\usepackage{subcaption}

\usepackage{epstopdf}

\DeclareGraphicsExtensions{.eps}

\newcommand{\highlight}[1]{\colorbox{cyan!20}{#1}}

\title{Exploring Domain Robust Lightweight Reward Models based on \\Router Mechanism}

\author{Hyuk Namgoong$^{1}$, Jeesu Jung$^{1}$, Sangkeun Jung$^{1}$\thanks{Corresponding author} \and Yoonhyung Roh$^{2}$ \\
  $^{1}$Computer Science and Engineering, Chungnam National University, Republic of Korea \\ $^{2}$Electronics and Telecommunications Research Institute, Republic of Korea \\
  \texttt{\{hyuk199, jisu.jung5, hugmanskj\}@gmail.com} \and \texttt{yhroh@etri.re.kr}
  }
\begin{document}
\maketitle
\begin{abstract}
Recent advancements in large language models have heavily relied on the large reward model from reinforcement learning from human feedback for fine-tuning. However, the use of a single reward model across various domains may not always be optimal, often requiring retraining from scratch when new domain data is introduced. To address these challenges, we explore the utilization of small language models operating in a domain-specific manner based on router mechanisms. Our three approaches are: 1) utilize mixture of experts to form a single reward model by modularizing an internal router and experts, 2) employing external router to select the appropriate reward model from multiple domain-specific models, and 3) the framework reduces parameter size by loading reward models and router adapters onto a single small language model using adapters. Experimental validation underscores the effectiveness of our approach, demonstrating performance comparable to baseline methods while also reducing the total parameter size.
\end{abstract}

\section{Introduction}
Most widely adopted Large Language Models (LLMs) have used the reward model of Reinforcement Learning from Human Feedback (RLHF) \cite{RLHF} for fine-tuning. These reward models are trained from various human feedback domains and are subsequently utilized as evaluation metrics during LLM fine-tuning processes.

However, training a single reward model across various domains to serve multiple purposes may lead to situations where the model is not fit for specific domains. Additionally, there is a challenge of retraining the reward model from scratch when new dataset from a new domain is introduced.

In this paper, we explore various router methods to address these challenges, as summarized in Table \ref{tab:methods}. Our approach, \textbf{M}ixture of \textbf{R}eward \textbf{E}xperts (MoRE), involves modularizing an internal router and experts within small language models to form a single reward model. \textbf{R}outer for \textbf{DO}main-specific reward model\textbf{S} (RODOS) employs an external router to select the appropriate reward model from multiple domain-specific reward models. The \textbf{A}dapter \textbf{R}outer \textbf{L}ightweight \textbf{I}ntegrated reward\textbf{S} \textbf{S}witching (ARLISS) framework applies adapters to load reward models and router adapters onto a single small language model, thereby reducing the parameter size of the multi-models.


To validate our methodologies, we conducted experiments with five different domains of reward datasets. In this experiment, our methods generally outperform the baseline, while RODOS shows the best performance. MoRE showcases a size reduction of about 52\%, while ARLISS achieves a reduction of approximately 55\% compared to the baseline.


\begin{table}[]
\resizebox{\columnwidth}{!}{%
\begin{tabular}{c|cccc}
\hline
\multirow{2}{*}{\textbf{Method}} & \multirow{2}{*}{\textbf{Router}}  & \multirow{2}{*}{\textbf{\makecell{Reward\\ model type}}} & \multicolumn{2}{c}{\textbf{Training}} \\ \cline{4-5}
& & & \textbf{Type}  & \textbf{Parameter} \\ \hline
Baseline                         & $\times$                          & Single           & Full             & All                \\ 
Base$_{LoRA}$                         & $\times$                          & Single           & \makecell{PEFT (LoRA)} & Partial      \\ 
\hline
MoRE                             & \makecell{$\ocircle$ (Internal)}  & Single           & Full             & All                \\ 
RODOS                            & \makecell{$\ocircle$ (External)}  & Multiple         & Full             & All                \\ 
ARLISS                           & \makecell{$\ocircle$ (External)}  & Multiple         & \makecell{PEFT (LoRA)} & Partial \\ \hline
\end{tabular}%
}
\caption{
Comparison of each method for the reward model. Baseline consists of a single reward model without a router. Base$_{LoRA}$ is similar to the baseline but applies Parameter-Efficient Fine-Tuning (PEFT) during training. Mixture of Reward Experts (MoRE) features an internal router but remains a single reward model. Router for DOmain-spcific reward modelS (RODOS) combines multiple reward models with a external router structure. Adapter Router Lightweight Integrated rewardS Switching (ARLISS) framework drastically reduces parameter size by applying PEFT to multiple reward models and external router.
}
\label{tab:methods}
\end{table}

\section{Related Works}

\begin{figure*}[!ht]
    \centering
    \begin{subfigure}[b]{0.182\textwidth}
        \includegraphics[width=\textwidth]{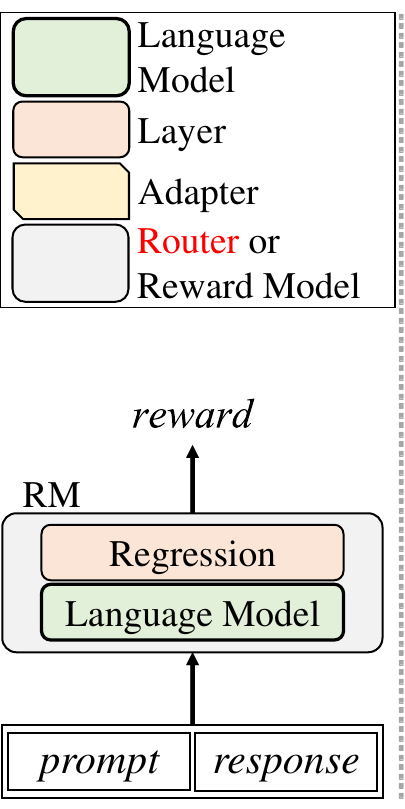}
        \caption{Baseline}
        \label{fig:Baseline}
    \end{subfigure}
    \begin{subfigure}[b]{0.25\textwidth}
        \includegraphics[width=\textwidth]{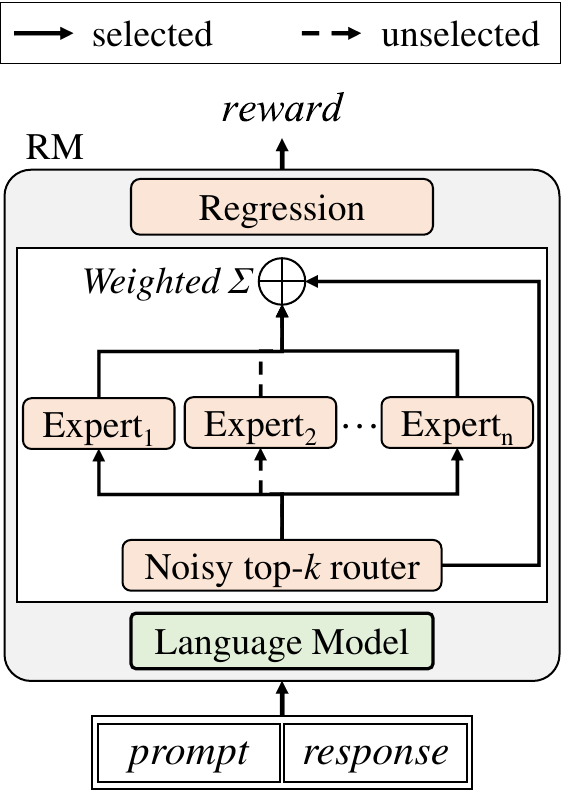}
        \caption{MoRE}
        \label{fig:MoRE}
    \end{subfigure}
    \begin{subfigure}[b]{0.242\textwidth}
        \includegraphics[width=\textwidth]{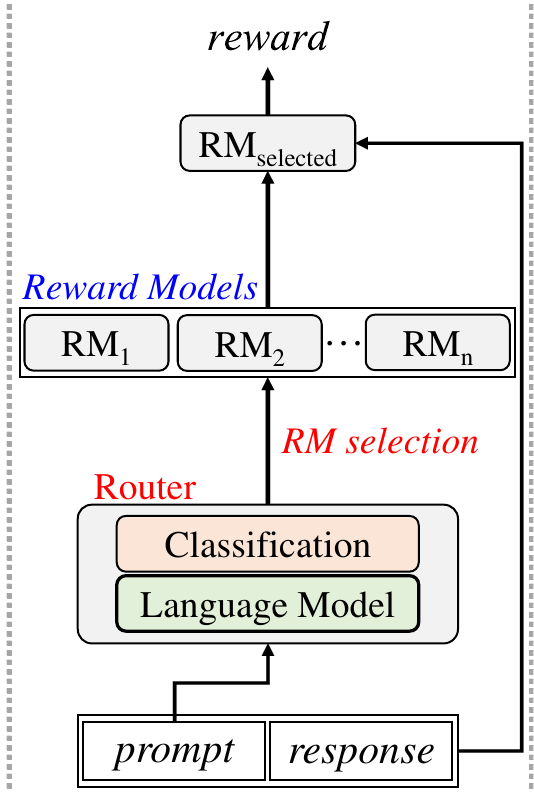}
        \caption{RODOS}
        \label{fig:RODOS}
    \end{subfigure}
    \begin{subfigure}[b]{0.26\textwidth}
        \includegraphics[width=\textwidth]{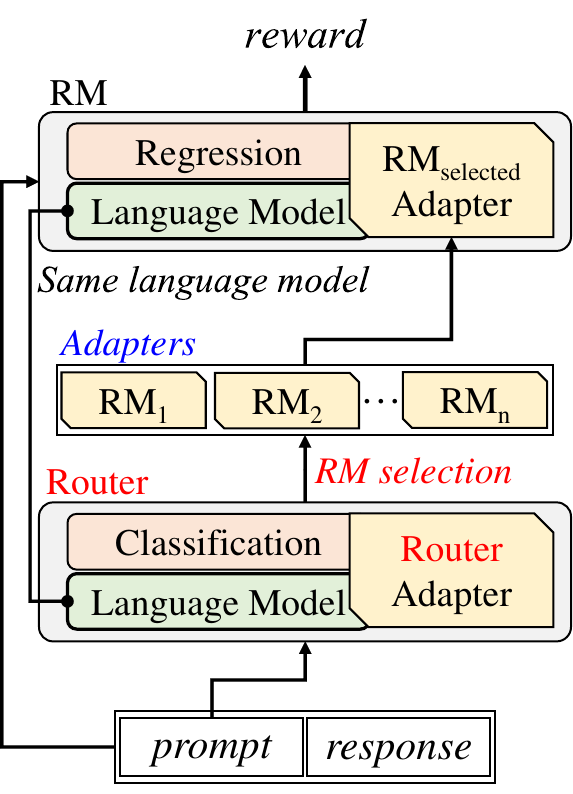}
        \caption{ARLISS}
        \label{fig:ARLISS}
    \end{subfigure}
    \caption{Illustration of each method. RM represents the reward model, and reward is scalar. $n$ denotes the number of domains, and Mixture of Reward Experts (MoRE) refers to Sparse Mixture of Experts with $k$ equal to 2. Router for DOmain-specific reward modelS (RODOS) involves loading all models for use, while the Adapter Router Lightweight Integrated rewardS Switching (ARLISS) framework loads only router and reward model adapters and a single language model, using adapter switching within the same language model.}
    \label{fig:overall}
\end{figure*}

Recent research focuses on improving LLMs \cite{palm, pythia, llama} training efficiency. Introducing the reward model serves to evaluate LLM performance in the RLHF fine-tuning method. In \cite{RLHF}, the reward model spans various domains, while \cite{black2023ddpo} applies the RLHF method to image generation. 

Research has explored methods for routing language models, such as routing LLMs \cite{shnitzer2023large,liu-liu-2021-simcls,ravaut-etal-2022-summareranker,jiang-etal-2023-llm}. Furthermore, various studies are underway to modularize and utilize routers within models\cite{mixtral,dikkala-etal-2023-benefits,peng2023efficient}, with Mixture of Experts (MoE) \cite{moe} being one. 

Research on efficient fine-tuning of language models is ongoing. Low-Rank Adaptation (LoRA) \cite{hu2021lora} attaches adapters to each layer and updates only the adapter parameters, enabling efficient learning. Building upon LoRA, further research explores efficiency improvements\cite{dettmers2023qlora, rajabzadehqdylora, babakniya2023slora} and additional tasks\cite{zhang-etal-2023-machine,Everaert_2023_ICCV,blattmann2023stable}. 

We do not train a single reward model across diverse domains. 
Instead, we utilize adapters to construct multi-reward models and routers, employing a small language model with LoRA, thereby reducing training time and parameters.

\section{Router Based Switching Reward Models}

The reward model assigns rewards to prompt and response. In RLHF, the reward model's loss function calculates the difference between the rewards for the chosen and rejected responses. Reward model dataset has the structure of one input prompt and least two of responses.

These reward models cover diverse domains like human preferences and toxic responses, using large-scale models. However, relying solely on one large model may not suit specific domains, and training from scratch for new domains takes time.

\subsection{Mixture of Reward Experts}
MoRE operates by having an internal router select suitable experts among several options, with both the router and experts modularized internally within the model. To implement MoRE, we utilize sparse MoE\cite{shazeer2017outrageously}, applying to small language models to create a single reward model. Maintaining the structure of a single reward model, it processes all dataset together during training, ensuring a training process similar to traditional method.

Sparse MoE, as depicted in Figure \ref{fig:MoRE}, utilizes noisy top-$k$ gating within the router layer directs the output to multiple expert layers before reaching the output layer. These expert layers follow a feed-forward network structure, computing a weighted sum based on the top-$k$ expert outputs, and then the regression layer generates the reward. Additionally, layer normalization is applied before the sparse MoE and regression layer.

\subsection{Router for Domain-Specific Reward Models}  

We introduce RODOS, in Figure \ref{fig:RODOS}, which involves training a small language model for each domain to create multiple domain-specific reward models. The external router is trained to select the reward model suitable for each prompt's domain. This resolves the challenge of a single large reward model trained across multiple domains, which may not be suitable for specific domains. 

Furthermore, RODOS offers a time-efficient solution by training new reward models for new data and retraining the router, rather than restarting the entire reward model training process. This efficiency is attributed to smaller model sizes and shorter router training times relative to reward model training.


\subsection{Adapter Router Lightweight Integrated Rewards Switching Framework}

Deploying all reward models and router creates deployment challenges for GPU memory. Hosting various models simultaneously results in the total parameter count becoming a multiple of the model parameters, thus demanding a considerable amount of GPU memory.

In the ARLISS framework, in Figure \ref{fig:ARLISS}, all reward models and routers are trained using adapters, with only the adapter parameters retained, and adapters are dynamically switched during inference. The router adapter selects and switches to the appropriate reward adapter during utilization. This approach consolidates multiple reward models and router into a single language model with multiple adapters, thereby reducing the total size of model parameters, making them lightweight.


We utilize Parameter-Efficient Fine-Tuning \cite{peft} alongside LoRA, functioning as the adapter mechanism. This enables efficient fine-tuning by updating only adapter parameters, contributing to the overall efficiency of the ARLISS framework.


\section{Experiments Setup}
\begin{table*}[!ht]
\centering
\resizebox{\textwidth}{!}{%
\begin{tabular}{clrcccccc}
\hline
\multirow{2}{*}{\textbf{Method}} & \multicolumn{1}{c}{\multirow{2}{*}{\textbf{\shortstack{Language\\ model}}}} &\multicolumn{1}{c}{\multirow{2}{*}{\textbf{\shortstack{Total\\ Parameter (M)}}}} & \multicolumn{6}{c}{\textbf{Accuracy}} \\ \cline{4-9}
& & & \textbf{Anthropic}   &\textbf{SHP}   &\textbf{HellaSwag} &\textbf{Dahoas}   &\textbf{Oasst}   &\textbf{Average}  \\ \hline
\multirow{2}{*}{Baseline}  & DeB$_{large}$  & 435\hspace{43pt}   & \textbf{0.6359}{\small.0058} & 0.6350 {\small.0117} & 0.4992 {\small.0009} & 0.9984 {\small.0003} & 0.7174 {\small.0053} & 0.6972  {\small.0048} \\
  & DeB$_{base}$  & 185 \hspace{4pt}(42.5\%)   & 0.6204 {\small.0031} & 0.6229 {\small.0054} & 0.5019 {\small.0025} & 0.9978 {\small.0008} & 0.7311 {\small.0060} & 0.6948  {\small.0036} \\

Base$_{LoRA}$  & DeB$_{base}$  & 187 \hspace{4pt}(43.0\%) & 0.6146 {\small.0053} & 0.6236 {\small.0083} & 0.4978 {\small.0012} & 0.9974 {\small.0007} & 0.7234 {\small.0072} & 0.6914  {\small.0030} \\ 
\hline
                                
\multirow{3}{*}{MoRE}     & DeB$_{base}$   & 207 \hspace{4pt}(47.6\%) & \highlight{0.6205}{\small.0032} & \highlight{0.6265}{\small.0080} & 0.4995          {\small.0010} & \highlight{0.9972}{\small.0009} & \highlight{\textbf{0.7368}}{\small.0099}  & \highlight{0.6961}{\small.0029} \\
                         & DeB$_{small}$  & 164  \hspace{4pt}(37.7\%) & 0.6097          {\small.0021} & 0.6187          {\small.0044} & 0.4944          {\small.0026} & 0.9965          {\small.0007} & 0.7180                      {\small.0089} & 0.6875          {\small.0024} \\
                         & DeB$_{xsmall}$ & 77   \hspace{4pt}(17.7\%) & 0.5892          {\small.0041} & 0.6117          {\small.0020} & \highlight{0.5019}{\small.0027} & 0.9945          {\small.0007} & 0.7207                      {\small.0023} & 0.6836          {\small.0015} \\ \hline

\multirow{3}{*}{RODOS}   & DeB$_{base}$   & 1,110 (255.2\%) & \highlight{0.6332}{\small.0005} & \highlight{0.6424}{\small.0017} & 0.4975                     {\small.0027} & \highlight{\textbf{0.9987}}{\small.0000}  & \highlight{0.7299} {\small.0002} & \highlight{\textbf{0.7003}}{\small.0007} \\
                         & DeB$_{small}$  & 846   (194.5\%) & 0.6236          {\small.0004} & 0.6367          {\small.0026} & 0.4969                     {\small.0027} & 0.9981                      {\small.0000} & 0.7290          {\small.0002} & 0.6969 {\small.0004} \\
                         & DeB$_{xsmall}$ & 420   \hspace{4pt}(96.6\%) & 0.5927          {\small.0002} & 0.6301          {\small.0023} & \highlight{\textbf{0.5072}}{\small.0027} & 0.9965                      {\small.0000} & 0.6961          {\small.0003} & 0.6845 {\small.0005} \\ \hline
                                
\multirow{3}{*}{ARLISS}  & DeB$_{base}$   & 197  \hspace{4pt}(45.3\%) & \highlight{0.6254}{\small.0004} & \highlight{\textbf{0.6525}}{\small.0017} & 0.4967          {\small.0000} & 0.9977                      {\small.0003} & 0.7150          {\small.0031} & \highlight{0.6975}{\small.0009} \\
                         & DeB$_{small}$  & 147  \hspace{4pt}(33.8\%) & 0.6167          {\small.0004} & 0.6297                     {\small.0010} & 0.4991          {\small.0001} & \highlight{0.9984}{\small.0005}              & \highlight{0.7240}{\small.0023} & 0.6936          {\small.0007} \\
                         & DeB$_{xsmall}$ & 76   \hspace{4pt}(17.5\%) & 0.6042          {\small.0004} & 0.6430                     {\small.0032} & \highlight{0.5018}{\small.0001} & 0.9975                      {\small.0001} & 0.7168          {\small.0050} & 0.6927          {\small.0007} \\ \hline
\end{tabular}
}
\caption{
Average performance across five domains and the total model parameters for each method. Language models are organized by the DeBERTaV3 (DeB) size. \highlight{Cyan highlight} indicates the best performance per our method within each domain, while \textbf{Bold} denotes the best performance across all methods. The parentheses in "Total Parameters" represent the percentage relative to the baseline size. Performances are evaluated with five seeds, and {\small $small\ numbers$} denotes standard deviation. 
}
\label{tab:result}
\end{table*}

\begin{table*}[!ht]
\centering
\resizebox{\textwidth}{!}{%
\begin{tabular}{clrccccccc}
\hline
\multirow{2}{*}{\textbf{Method}} & \multicolumn{1}{c}{\multirow{2}{*}{\textbf{\begin{tabular}[c]{@{}c@{}}Language\\ model\end{tabular}}}} & \multicolumn{1}{c}{\multirow{2}{*}{\textbf{\shortstack{Total\\ Parameter (M)}}}} & \multicolumn{7}{c}{\textbf{1 epoch train time (sec)}} \\ \cline{4-10} 
& & & \textbf{Total} & \textbf{Anthropic} & \textbf{SHP} & \textbf{HellaSwag} & \textbf{Dahoas} & \textbf{Oasst} & \textbf{Router} \\ \hline
Baseline               & DeB$_{large}$   \hspace{2pt}(435) & 435\hspace{43pt}   & 17,392    & -     & - & - & - & - & - \\ \hline
\multirow{1}{*}{MoRE}  & DeB$_{base}$    \hspace{6pt}(184) & 207 \hspace{4pt}(47.6$\%$)   & 5,010     & -     & - & - & - & - & - \\
\multirow{1}{*}{RODOS} & DeB$_{base}$    \hspace{6pt}(184) & 1,110 (255.2$\%$) & 7,355     & 2,768 & 696   & 702   & 528   & 235   & 2,426 \\
\multirow{1}{*}{ARLISS}& DeB$_{base}$    \hspace{6pt}(184) & 197 \hspace{4pt}(45.3$\%$)   & 7,067     & 2,682 & 663   & 672   & 506   & 216   & 2,328 \\
                       \hline
\end{tabular}%
}
\caption{
Training time for 1 epoch and the total model parameters for each method. Language models are DeBERTaV3 (DeB) used in the experiments, with the original parameter size(M) indicated in parentheses. Baseline and Mixture of Reward Experts (MoRE) are single models, so only the total time is presented. The parentheses in "Total Parameters" represent the percentage relative to the baseline.
}
\label{tab:time}
\end{table*}

\subsection{Datasets}
In this study, we validate the methodology using reward model datasets from five different domains. In cases where the dataset structure is unsuitable for training a reward model, we convert it to a suitable reward dataset structure using only English data. 

Anthropic dataset detects toxic responses and distinguishes whether a response is helpful or harmless \cite{anthropic}.
SHP is a dataset that has two human-written summary responses in a given context \cite{shp}.
HellaSwag is a dataset used for sentence completion tasks, featuring multiple responses to a given prompt \cite{hellaswag}.
Dahoas is a dataset where the model generates two responses to a prompt and humans distinguish between good and bad responses \cite{dahoas}. 
Oasst is a dataset that has ranked human-written responses in a given prompt \footnote{\url{https://huggingface.co/datasets/OpenAssistant/oasst2}}.The conversion of each dataset into a reward dataset structure is detailed in Appendix \ref{sec:convert}

\subsection{Language Models}\label{sec:model}
We employed the encoder-only model DeBERTaV3(DeB) \cite{debertav3}, which leverages Transformer's encoder. 
For our methods, we implement language models such as DeB$_{base}$, DeB$_{small}$, and DeB$_{xsmall}$. The router model is implement with the same language model as the reward model.

\subsection{Baseline Methods}\label{sec:base}
 In Table \ref{tab:methods}, the baseline method is a traditional single reward model trained without a router. This method is implemented using DeB$_{large}$ and DeB$_{base}$ for comparison with our proposed approaches. During fine-tuning, all datasets are processed together. Preliminary experiments with other models are detailed in the Appendix \ref{sec:preliminary}. 

Additionally, Base$_{LoRA}$ was included in the experiments. This method follows the same training process as the baseline but incorporates LoRA. The purpose is to determine if applying LoRA yields higher performance than the baseline DeB$_{large}$. However, it was observed that Base$_{LoRA}$ exhibited lower performance. Base$_{LoRA}$ were conducted using DeB${base}$.

\subsection{Evaluation Metric for Reward Model}\label{sec:reward_model}
To evaluate the performance of reward model, we utilized $binary\ accuracy$. During reward computation for each prompt-response pair, if the reward for the chosen response exceeds that of the rejected response, it is classified as $true$; otherwise, it is classified as $false$.

\section{Experimental Results}

Our study investigates the effectiveness of the proposed router methods through experimental analyses, focusing on key aspects: evaluating the router's impact on application performance, analyzing training time across methods, and comparing total parameters with and without ARLISS integration.

\subsection{Reward Models Performance}\label{sec:analyzing-performance}
We analyze the accuracy of our proposed framework compared to other methods. In this regard, we conduct statistical significance analysis for each test dataset. To ensure meaningful evaluation, we conduct evaluations with 5 different seeds.

Table \ref{tab:result} displays the accuracy for each dataset's test data and the corresponding average. Generally, when the accuracy is less than 0.02, it is considered statistically similar. 
Excluding the Anthropic dataset, our methods generally outperform the baseline, with RODOS showing the best performance. Moreover, MoRE and ARLISS demonstrate a size reduction of approximately half of the baseline. This suggests that our methods offer the potential to replace the baseline with smaller model sizes.

\subsection{Training Time}\label{sec:analyzing-time}
We analyze the implementation time for models with and without the router. For multi-reward model methods, we assess the training time for each reward model and the router. For single reward model methods, only the training time for the reward model across all datasets is considered.

Table \ref{tab:time} presents the training time for each method per epoch. Overall, our methods show a reduction in time by approximately 63\%, with ARLISS demonstrating around a 5\% decrease compared to RODOS.


\subsection{Total Parameter Size}\label{sec:analyzing-parameters}
We analyze our ARLISS framework along with other methods. In this context, we perform parameter size analysis using the same language model.

Table \ref{tab:time} reveals that the ARLISS framework boasts the smallest parameter size. MoRE showcases a size reduction of about 52\%, while ARLISS achieves a reduction of approximately 55\% compared to the baseline. Although ARLISS employs a multi-reward model structure, it features 10 million fewer parameters than MoRE and achieves over an 80\% reduction compared to RODOS, another multi-reward model framework.

\section{Conclusion}
In addressing the limitations of a single large reward model, which can be unsuitable for specific domains and requires retraining when new domain data is introduced, we have implemented router methods. MoRE features an internal router alongside a single small reward model, while RODOS incorporates an external router and domain-specific reward models. These methods effectively mitigate challenges related to domain specificity and the need for retraining when new domain data is introduced. Moreover, the ARLISS framework, with adapters for routers and multi-reward models, shows potential for GPU memory optimization by reducing model size.

Further research will focus on optimizing the ARLISS framework. Additionally, we plan to investigate the integration of the ARLISS framework into MoRE.

\section*{Limitation}
The ARLISS framework requires more inference time compared to RODOS, as discussed in Appendix \ref{sec:appinference}. This delay arises from the router selecting the reward model and switching the adapter within the same language model, resulting in time consumption during the switching process.

\section*{Acknowledgements}
This work was supported by 
Institute of Information \& Communications Technology Planning \& Evaluation(IITP) grant funded by the Korea government(MSIT) (2019-0-00004, Development of semi-supervised learning language intelligence technology and Korean tutoring service for foreigners),
the National Research Foundation of Korea(NRF) grant funded by the Korea government(MSIT)(No. 2022R1F1A1071047)
 and 
research fund of Chungnam National University.

\bibliography{anthology,custom}

\begin{thebibliography}{32}
\expandafter\ifx\csname natexlab\endcsname\relax\def\natexlab#1{#1}\fi

\bibitem[{{Alex Havrilla}(2023)}]{dahoas}
{Alex Havrilla}. 2023.
\newblock \href {https://doi.org/10.57967/hf/1428} {synthetic-instruct-gptj-pairwise (revision cc92d8d)}.

\bibitem[{Babakniya et~al.(2023)Babakniya, Elkordy, Ezzeldin, Liu, Song, El-Khamy, and Avestimehr}]{babakniya2023slora}
Sara Babakniya, Ahmed~Roushdy Elkordy, Yahya~H Ezzeldin, Qingfeng Liu, Kee-Bong Song, Mostafa El-Khamy, and Salman Avestimehr. 2023.
\newblock Slora: Federated parameter efficient fine-tuning of language models.
\newblock \emph{arXiv preprint arXiv:2308.06522}.

\bibitem[{Bai et~al.(2022)Bai, Jones, Ndousse, Askell, Chen, DasSarma, Drain, Fort, Ganguli, Henighan, Joseph, Kadavath, Kernion, Conerly, El-Showk, Elhage, Hatfield-Dodds, Hernandez, Hume, Johnston, Kravec, Lovitt, Nanda, Olsson, Amodei, Brown, Clark, McCandlish, Olah, Mann, and Kaplan}]{anthropic}
Yuntao Bai, Andy Jones, Kamal Ndousse, Amanda Askell, Anna Chen, Nova DasSarma, Dawn Drain, Stanislav Fort, Deep Ganguli, Tom Henighan, Nicholas Joseph, Saurav Kadavath, Jackson Kernion, Tom Conerly, Sheer El-Showk, Nelson Elhage, Zac Hatfield-Dodds, Danny Hernandez, Tristan Hume, Scott Johnston, Shauna Kravec, Liane Lovitt, Neel Nanda, Catherine Olsson, Dario Amodei, Tom Brown, Jack Clark, Sam McCandlish, Chris Olah, Ben Mann, and Jared Kaplan. 2022.
\newblock \href {http://arxiv.org/abs/2204.05862} {Training a helpful and harmless assistant with reinforcement learning from human feedback}.

\bibitem[{Bhargava et~al.(2021)Bhargava, Drozd, and Rogers}]{bert_small1}
Prajjwal Bhargava, Aleksandr Drozd, and Anna Rogers. 2021.
\newblock \href {http://arxiv.org/abs/2110.01518} {Generalization in nli: Ways (not) to go beyond simple heuristics}.

\bibitem[{Biderman et~al.(2023)Biderman, Schoelkopf, Anthony, Bradley, O'Brien, Hallahan, Khan, Purohit, Prashanth, Raff, Skowron, Sutawika, and van~der Wal}]{pythia}
Stella Biderman, Hailey Schoelkopf, Quentin Anthony, Herbie Bradley, Kyle O'Brien, Eric Hallahan, Mohammad~Aflah Khan, Shivanshu Purohit, USVSN~Sai Prashanth, Edward Raff, Aviya Skowron, Lintang Sutawika, and Oskar van~der Wal. 2023.
\newblock \href {http://arxiv.org/abs/2304.01373} {Pythia: A suite for analyzing large language models across training and scaling}.

\bibitem[{Black et~al.(2023)Black, Janner, Du, Kostrikov, and Levine}]{black2023ddpo}
Kevin Black, Michael Janner, Yilun Du, Ilya Kostrikov, and Sergey Levine. 2023.
\newblock \href {http://arxiv.org/abs/2305.13301} {Training diffusion models with reinforcement learning}.

\bibitem[{Blattmann et~al.(2023)Blattmann, Dockhorn, Kulal, Mendelevitch, Kilian, Lorenz, Levi, English, Voleti, Letts, Jampani, and Rombach}]{blattmann2023stable}
Andreas Blattmann, Tim Dockhorn, Sumith Kulal, Daniel Mendelevitch, Maciej Kilian, Dominik Lorenz, Yam Levi, Zion English, Vikram Voleti, Adam Letts, Varun Jampani, and Robin Rombach. 2023.
\newblock \href {http://arxiv.org/abs/2311.15127} {Stable video diffusion: Scaling latent video diffusion models to large datasets}.

\bibitem[{Chen et~al.(2022)Chen, Deng, Wu, Gu, and Li}]{moe}
Zixiang Chen, Yihe Deng, Yue Wu, Quanquan Gu, and Yuanzhi Li. 2022.
\newblock \href {http://arxiv.org/abs/2208.02813} {Towards understanding mixture of experts in deep learning}.

\bibitem[{Chowdhery et~al.(2022)Chowdhery, Narang, Devlin, Bosma, Mishra, Roberts, Barham, Chung, Sutton, Gehrmann et~al.}]{palm}
Aakanksha Chowdhery, Sharan Narang, Jacob Devlin, Maarten Bosma, Gaurav Mishra, Adam Roberts, Paul Barham, Hyung~Won Chung, Charles Sutton, Sebastian Gehrmann, et~al. 2022.
\newblock Palm: Scaling language modeling with pathways.
\newblock \emph{arXiv preprint arXiv:2204.02311}.

\bibitem[{Dettmers et~al.(2023)Dettmers, Pagnoni, Holtzman, and Zettlemoyer}]{dettmers2023qlora}
Tim Dettmers, Artidoro Pagnoni, Ari Holtzman, and Luke Zettlemoyer. 2023.
\newblock \href {http://arxiv.org/abs/2305.14314} {Qlora: Efficient finetuning of quantized llms}.

\bibitem[{Devlin et~al.(2018)Devlin, Chang, Lee, and Toutanova}]{bert}
Jacob Devlin, Ming{-}Wei Chang, Kenton Lee, and Kristina Toutanova. 2018.
\newblock \href {http://arxiv.org/abs/1810.04805} {{BERT:} pre-training of deep bidirectional transformers for language understanding}.
\newblock \emph{CoRR}, abs/1810.04805.

\bibitem[{Dikkala et~al.(2023)Dikkala, Ghosh, Meka, Panigrahy, Vyas, and Wang}]{dikkala-etal-2023-benefits}
Nishanth Dikkala, Nikhil Ghosh, Raghu Meka, Rina Panigrahy, Nikhil Vyas, and Xin Wang. 2023.
\newblock \href {https://doi.org/10.18653/v1/2023.emnlp-main.583} {On the benefits of learning to route in mixture-of-experts models}.
\newblock In \emph{Proceedings of the 2023 Conference on Empirical Methods in Natural Language Processing}, pages 9376--9396, Singapore. Association for Computational Linguistics.

\bibitem[{Ethayarajh et~al.(2022)Ethayarajh, Choi, and Swayamdipta}]{shp}
Kawin Ethayarajh, Yejin Choi, and Swabha Swayamdipta. 2022.
\newblock \href {https://proceedings.mlr.press/v162/ethayarajh22a.html} {Understanding dataset difficulty with $\mathcal{V}$-usable information}.
\newblock In \emph{Proceedings of the 39th International Conference on Machine Learning}, volume 162 of \emph{Proceedings of Machine Learning Research}, pages 5988--6008. PMLR.

\bibitem[{Everaert et~al.(2023)Everaert, Bocchio, Arpa, S\"usstrunk, and Achanta}]{Everaert_2023_ICCV}
Martin~Nicolas Everaert, Marco Bocchio, Sami Arpa, Sabine S\"usstrunk, and Radhakrishna Achanta. 2023.
\newblock Diffusion in style.
\newblock In \emph{Proceedings of the IEEE/CVF International Conference on Computer Vision (ICCV)}, pages 2251--2261.

\bibitem[{He et~al.(2021)He, Gao, and Chen}]{debertav3}
Pengcheng He, Jianfeng Gao, and Weizhu Chen. 2021.
\newblock \href {http://arxiv.org/abs/2111.09543} {Debertav3: Improving deberta using electra-style pre-training with gradient-disentangled embedding sharing}.

\bibitem[{Hu et~al.(2021)Hu, Shen, Wallis, Allen-Zhu, Li, Wang, Wang, and Chen}]{hu2021lora}
Edward~J. Hu, Yelong Shen, Phillip Wallis, Zeyuan Allen-Zhu, Yuanzhi Li, Shean Wang, Lu~Wang, and Weizhu Chen. 2021.
\newblock \href {http://arxiv.org/abs/2106.09685} {Lora: Low-rank adaptation of large language models}.

\bibitem[{Jiang et~al.(2024)Jiang, Sablayrolles, Roux, Mensch, Savary, Bamford, Chaplot, de~las Casas, Hanna, Bressand, Lengyel, Bour, Lample, Lavaud, Saulnier, Lachaux, Stock, Subramanian, Yang, Antoniak, Scao, Gervet, Lavril, Wang, Lacroix, and Sayed}]{mixtral}
Albert~Q. Jiang, Alexandre Sablayrolles, Antoine Roux, Arthur Mensch, Blanche Savary, Chris Bamford, Devendra~Singh Chaplot, Diego de~las Casas, Emma~Bou Hanna, Florian Bressand, Gianna Lengyel, Guillaume Bour, Guillaume Lample, Lélio~Renard Lavaud, Lucile Saulnier, Marie-Anne Lachaux, Pierre Stock, Sandeep Subramanian, Sophia Yang, Szymon Antoniak, Teven~Le Scao, Théophile Gervet, Thibaut Lavril, Thomas Wang, Timothée Lacroix, and William~El Sayed. 2024.
\newblock \href {http://arxiv.org/abs/2401.04088} {Mixtral of experts}.

\bibitem[{Jiang et~al.(2023)Jiang, Ren, and Lin}]{jiang-etal-2023-llm}
Dongfu Jiang, Xiang Ren, and Bill~Yuchen Lin. 2023.
\newblock \href {https://doi.org/10.18653/v1/2023.acl-long.792} {{LLM}-blender: Ensembling large language models with pairwise ranking and generative fusion}.
\newblock In \emph{Proceedings of the 61st Annual Meeting of the Association for Computational Linguistics (Volume 1: Long Papers)}, pages 14165--14178, Toronto, Canada. Association for Computational Linguistics.

\bibitem[{Liu et~al.(2019)Liu, Ott, Goyal, Du, Joshi, Chen, Levy, Lewis, Zettlemoyer, and Stoyanov}]{roberta}
Yinhan Liu, Myle Ott, Naman Goyal, Jingfei Du, Mandar Joshi, Danqi Chen, Omer Levy, Mike Lewis, Luke Zettlemoyer, and Veselin Stoyanov. 2019.
\newblock \href {http://arxiv.org/abs/1907.11692} {Roberta: {A} robustly optimized {BERT} pretraining approach}.
\newblock \emph{CoRR}, abs/1907.11692.

\bibitem[{Liu and Liu(2021)}]{liu-liu-2021-simcls}
Yixin Liu and Pengfei Liu. 2021.
\newblock \href {https://doi.org/10.18653/v1/2021.acl-short.135} {{S}im{CLS}: A simple framework for contrastive learning of abstractive summarization}.
\newblock In \emph{Proceedings of the 59th Annual Meeting of the Association for Computational Linguistics and the 11th International Joint Conference on Natural Language Processing (Volume 2: Short Papers)}, pages 1065--1072, Online. Association for Computational Linguistics.

\bibitem[{Mangrulkar et~al.(2022)Mangrulkar, Gugger, Debut, Belkada, Paul, and Bossan}]{peft}
Sourab Mangrulkar, Sylvain Gugger, Lysandre Debut, Younes Belkada, Sayak Paul, and Benjamin Bossan. 2022.
\newblock Peft: State-of-the-art parameter-efficient fine-tuning methods.
\newblock \url{https://github.com/huggingface/peft}.

\bibitem[{Ouyang et~al.(2022)Ouyang, Wu, Jiang, Almeida, Wainwright, Mishkin, Zhang, Agarwal, Slama, Ray et~al.}]{RLHF}
Long Ouyang, Jeffrey Wu, Xu~Jiang, Diogo Almeida, Carroll Wainwright, Pamela Mishkin, Chong Zhang, Sandhini Agarwal, Katarina Slama, Alex Ray, et~al. 2022.
\newblock Training language models to follow instructions with human feedback.
\newblock \emph{Advances in Neural Information Processing Systems}, 35:27730--27744.

\bibitem[{Peng et~al.(2023)Peng, Burns, Chen, Parthasarathy, and Ning}]{peng2023efficient}
Bo~Peng, Ben Burns, Ziqi Chen, Srinivasan Parthasarathy, and Xia Ning. 2023.
\newblock \href {http://arxiv.org/abs/2310.01612} {Towards efficient and effective adaptation of large language models for sequential recommendation}.

\bibitem[{Radford et~al.(2019)Radford, Wu, Child, Luan, Amodei, and Sutskever}]{gpt}
Alec Radford, Jeff Wu, Rewon Child, David Luan, Dario Amodei, and Ilya Sutskever. 2019.
\newblock Language models are unsupervised multitask learners.

\bibitem[{Rajabzadeh et~al.()Rajabzadeh, Valipour, Tahaei, Kwon, Ghodsi, Chen, and Rezagholizadeh}]{rajabzadehqdylora}
Hossein Rajabzadeh, Mojtaba Valipour, Marzieh Tahaei, Hyock~Ju Kwon, Ali Ghodsi, Boxing Chen, and Mehdi Rezagholizadeh.
\newblock Qdylora: Quantized dynamic low-rank adaptation for efficient large language model tuning.

\bibitem[{Ravaut et~al.(2022)Ravaut, Joty, and Chen}]{ravaut-etal-2022-summareranker}
Mathieu Ravaut, Shafiq Joty, and Nancy Chen. 2022.
\newblock \href {https://doi.org/10.18653/v1/2022.acl-long.309} {{S}umma{R}eranker: A multi-task mixture-of-experts re-ranking framework for abstractive summarization}.
\newblock In \emph{Proceedings of the 60th Annual Meeting of the Association for Computational Linguistics (Volume 1: Long Papers)}, pages 4504--4524, Dublin, Ireland. Association for Computational Linguistics.

\bibitem[{Shazeer et~al.(2017)Shazeer, Mirhoseini, Maziarz, Davis, Le, Hinton, and Dean}]{shazeer2017outrageously}
Noam Shazeer, Azalia Mirhoseini, Krzysztof Maziarz, Andy Davis, Quoc Le, Geoffrey Hinton, and Jeff Dean. 2017.
\newblock \href {http://arxiv.org/abs/1701.06538} {Outrageously large neural networks: The sparsely-gated mixture-of-experts layer}.

\bibitem[{Shnitzer et~al.(2023)Shnitzer, Ou, Silva, Soule, Sun, Solomon, Thompson, and Yurochkin}]{shnitzer2023large}
Tal Shnitzer, Anthony Ou, Mírian Silva, Kate Soule, Yuekai Sun, Justin Solomon, Neil Thompson, and Mikhail Yurochkin. 2023.
\newblock \href {http://arxiv.org/abs/2309.15789} {Large language model routing with benchmark datasets}.

\bibitem[{Touvron et~al.(2023)Touvron, Lavril, Izacard, Martinet, Lachaux, Lacroix, Rozière, Goyal, Hambro, Azhar, Rodriguez, Joulin, Grave, and Lample}]{llama}
Hugo Touvron, Thibaut Lavril, Gautier Izacard, Xavier Martinet, Marie-Anne Lachaux, Timothée Lacroix, Baptiste Rozière, Naman Goyal, Eric Hambro, Faisal Azhar, Aurelien Rodriguez, Armand Joulin, Edouard Grave, and Guillaume Lample. 2023.
\newblock \href {http://arxiv.org/abs/2302.13971} {Llama: Open and efficient foundation language models}.

\bibitem[{Turc et~al.(2019)Turc, Chang, Lee, and Toutanova}]{bert_small2}
Iulia Turc, Ming{-}Wei Chang, Kenton Lee, and Kristina Toutanova. 2019.
\newblock \href {http://arxiv.org/abs/1908.08962} {Well-read students learn better: The impact of student initialization on knowledge distillation}.
\newblock \emph{CoRR}, abs/1908.08962.

\bibitem[{Zellers et~al.(2019)Zellers, Holtzman, Bisk, Farhadi, and Choi}]{hellaswag}
Rowan Zellers, Ari Holtzman, Yonatan Bisk, Ali Farhadi, and Yejin Choi. 2019.
\newblock Hellaswag: Can a machine really finish your sentence?
\newblock In \emph{Proceedings of the 57th Annual Meeting of the Association for Computational Linguistics}.

\bibitem[{Zhang et~al.(2023)Zhang, Rajabi, Duh, and Koehn}]{zhang-etal-2023-machine}
Xuan Zhang, Navid Rajabi, Kevin Duh, and Philipp Koehn. 2023.
\newblock \href {https://doi.org/10.18653/v1/2023.wmt-1.43} {Machine translation with large language models: Prompting, few-shot learning, and fine-tuning with {QL}o{RA}}.
\newblock In \emph{Proceedings of the Eighth Conference on Machine Translation}, pages 468--481, Singapore. Association for Computational Linguistics.

\end{thebibliography}

\appendix

\section{Hyperparameter Settings}\label{sec:apphyperparameter}
In this section, we provide details of the hyperparameter and LoRA settings in our experiments.

Each model is trained with the same hyperparameters to evaluate under identical conditions. Training utilizes a learning rate of 5.0e-6, a batch size of 32, and 3 epochs, with the AdamW optimizer. However, DeB$_{large}$ is trained with batch size of 8 due to memory limitations.

For LoRA, we established the projection layer for $query$, $key$, and $value$, along with the $dense$ module. 
We set the rank to 12, alpha to 768, and dropout to 0.1 based on the layers and dimensions of DeB$_{base}$. The experiments were conducted using Nvidia V100 GPUs.

\section{Conversion to Reward Dataset Structure}\label{sec:convert}

In this section, we discuss the process of converting each dataset into the structure of a reward model dataset. First, we introduce the reward model dataset, which consists of one input prompt and at least two responses. Each response is designated as either chosen or rejected, and the reward model learns to assign higher reward to the chosen response compared to the rejected response when given the prompt and response as input. The requirement of "at least two responses" means that responses must be paired as \textit{chosen} and \textit{rejected}; if there are more than two responses, ranking or selecting is performed to pair them into sets.

Anthropic resembles the reward model dataset but combines the prompt and response. To facilitate the training of a reward model, we preprocess it by separating human input as the prompt and the Assistant's response as responses, resulting in a format of one prompt and two responses.

SHP consists of two human-written summary responses in a given context. Based on the desired human-written summary label, we select chosen and rejected responses for the context.

HellaSwag involves sentence completion tasks with more than two endings. We designate the correct endings as chosen and randomly select from the incorrect endings as rejected responses.

Dahoas and Oasst did not require separate conversion into reward model datasets. However, since our experiments were conducted in English, we only used English data from Oasst, which contains multiple languages.

\section{Size of Datasets}\label{sec:appsize}
In this section, Table \ref{tab:train_count} and \ref{tab:test_count} presents the sizes of the datasets used in the experiments. These datasets are used for train and test the reward models and router.
\begin{table}[!h]
\centering 
\begin{tabular}{ccc}
\hline
\textbf{Dataset} & \textbf{\# of data} & \textbf{\% of data} \\ \hline
Anthropic        & 80,307              & 57.02               \\
SHP              & 19,493              & 13.84               \\
HellaSwag        & 19,952              & 14.17               \\
Dahoas           & 14,913              & 10.59               \\
Oasst            & 6,176               & 4.39                \\ \hline
Total            & 140,841             & 100                 \\ \hline
\end{tabular}
\caption{
\label{tab:train_count} 
Data size used to train the reward model and router. The number of data for each domain as a percentage of the total training data. 
}
\end{table}

\begin{table}[!h]
\centering 
\begin{tabular}{ccc}
\hline
\textbf{Dataset} & \textbf{\# of data} & \textbf{\% of data} \\ \hline
Anthropic        & 8,539              & 34.59               \\
SHP              & 2,166              & 8.77                \\
HellaSwag        & 10,003             & 40.52               \\
Dahoas           & 3,313              & 13.42               \\
Oasst            & 668                & 2.71                \\ \hline
Total            & 24,689             & 100                 \\ \hline
\end{tabular}
\caption{
\label{tab:test_count} 
Data size used to test the reward model and router. The number of data for each domain as a percentage of the total testing data. 
}
\end{table}

\section{Inference Time}\label{sec:appinference}
In this section, Table \ref{tab:inference_time} provides the inference times for each method and language model. The experiments were conducted using a total of 2500 data samples. We measure the time it takes for the method to process one input from each dataset.
\begin{table}[!h]
\centering 
{\small
\begin{tabular}{clc}
\hline
\textbf{Method}           & \textbf{Language model} & \textbf{1step(sec)} \\ \hline
Baseline                  & DeB$_{large}$  & 0.08       \\ \hline
\multirow{3}{*}{MoRE}     & DeB$_{base}$   & 0.04       \\
                          & DeB$_{small}$  & 0.02       \\
                          & DeB$_{xsmall}$ & 0.04       \\ \hline
\multirow{3}{*}{RODOS}    & DeB$_{base}$   & 0.08       \\
                          & DeB$_{small}$  & 0.05       \\
                          & DeB$_{xsmall}$ & 0.08       \\ \hline
\multirow{3}{*}{ARLISS}    
                          & DeB$_{base}$   & 0.19       \\
                          & DeB$_{small}$  & 0.10       \\
                          & DeB$_{xsmall}$ & 0.19       \\ \hline
\end{tabular}%
}
\caption{
The inference time is measured for each method and language model. We select 500 samples from each of the five test datasets used in the experiment, measure the inference time, and calculate the average.
}
\label{tab:inference_time}
\end{table}

\section{Preliminary Model Selection Experiments}\label{sec:preliminary}
In this section, Table \ref{tab:ap_result} presents the results of preliminary experiments conducted to determine the models to be used in subsequent experiments.The Baseline method was applied using DeBERTaV3 and four other models: BERT$_{base}$ \cite{bert}, BERT$_{small}$ \cite{bert_small1, bert_small2}, RoBERTa$_{base}$ \cite{roberta}, and GPT-2 \cite{gpt}. These results help in assessing the performance and suitability of each model for the primary experiments. The experiments are conducted with five seeds each, and the performance metrics are averaged and standard deviation is computed accordingly.

\begin{table*}[!h]
\centering
\resizebox{\textwidth}{!}{%
\begin{tabular}{lccccccc}
\hline
\multicolumn{1}{c}{\multirow{2}{*}{\textbf{\shortstack{Language\\ model}}}} &\multicolumn{1}{c}{\multirow{2}{*}{\textbf{\shortstack{Parameter\\ Size (M)}}}} & \multicolumn{6}{c}{\textbf{Accuracy}} \\ \cline{3-8}
               &     & \textbf{Anthropic}   &\textbf{SHP}          &\textbf{HellaSwag}    &\textbf{Dahoas}       &\textbf{Oasst}        &\textbf{Average}       \\ 
\hline
DeB$_{large}$    & 435 & 0.6359 {\small.0058} & 0.6350 {\small.0117} & 0.4992 {\small.0009} & 0.9984 {\small.0003} & 0.7174 {\small.0053} & 0.6972  {\small.0048} \\
DeB$_{base}$     & 185 & 0.6204 {\small.0031} & 0.6229 {\small.0054} & 0.5019 {\small.0025} & 0.9978 {\small.0008} & 0.7311 {\small.0060} & 0.6948  {\small.0036} \\
DeB$_{small}$    & 141 & 0.6046 {\small.0052} & 0.6213 {\small.0035} & 0.4926 {\small.0021} & 0.9963 {\small.0011} & 0.7156 {\small.0097} & 0.6861  {\small.0043} \\
DeB$_{xsmall}$   & 70  & 0.5853 {\small.0051} & 0.6165 {\small.0061} & 0.5016 {\small.0016} & 0.9956 {\small.0007} & 0.7213 {\small.0022} & 0.6841  {\small.0031} \\
\hline
BERT$_{base}$    & 109 & 0.6157 {\small.0042} & 0.6095 {\small.0050} & 0.4993 {\small.0032} & 0.9951 {\small.0009} & 0.7087 {\small.0083} & 0.6857  {\small.0043} \\
BERT$_{small}$   & 29  & 0.5857 {\small.0032} & 0.6156 {\small.0070} & 0.4986 {\small.0020} & 0.9917 {\small.0011} & 0.7117 {\small.0080} & 0.6807  {\small.0043} \\
RoBERTa$_{base}$ & 125 & 0.6241 {\small.0029} & 0.6194 {\small.0058} & 0.4973 {\small.0009} & 0.9974 {\small.0008} & 0.7162 {\small.0126} & 0.6909  {\small.0046} \\
GPT-2          & 124 & 0.5987 {\small.0031} & 0.6206 {\small.0064} & 0.4954 {\small.0018} & 0.9925 {\small.0021} & 0.6904 {\small.0124} & 0.6795  {\small.0052} \\
\hline
\end{tabular}
}
\caption{
Average performance across five domains and model parameter sizes for experiments using the Baseline method. The language models include DeBERTaV3 (DeB) as used in the paper, BERT$_{base}$, BERT$_{small}$, RoBERTa$_{base}$, and GPT-2. Performances are evaluated with five seeds, and {\small $small\ numbers$} denotes standard deviation.
}
\label{tab:ap_result}
\end{table*}

\end{document}